\documentclass{article}

\PassOptionsToPackage{numbers, compress}{natbib}

\usepackage[final]{nips_2018}

\usepackage[utf8]{inputenc} 
\usepackage[T1]{fontenc}    
\usepackage{hyperref}       
\usepackage{url}            
\usepackage{booktabs}       
\usepackage{amsfonts}       
\usepackage{nicefrac}       
\usepackage{csquotes}       
\usepackage{float}
\usepackage{caption}
\usepackage{amsmath}
\usepackage{multicol}
\usepackage{graphicx}            

\title{
Three Tools for Practical Differential Privacy
}

\author{
  Koen Lennart van der Veen \\
  Graduate School of Informatics\\
  University of Amsterdam\\
  \texttt{koen.vanderveen@polis.global} \\
   \And
   Ruben Seggers \\
   Graduate School of Informatics\\
   University of Amsterdam\\
   \texttt{ruben.seggers@polis.global}   
   \And
   Peter Bloem\\
   KRR group\\
   VU Amsterdam\\
   \texttt{vu@peterbloem.nl}  \\
  \And
  Giorgio Patrini \\
  UvA Bosch Delta Lab \\
  University of Amsterdam\\
  \texttt{g.patrini@uva.nl }  \\
}

\begin{document}

\maketitle

\begin{abstract}

Differentially private learning on real-world data poses challenges for standard machine learning practice: privacy guarantees are difficult to interpret, hyperparameter tuning on private data reduces the privacy budget, and ad-hoc privacy attacks are often required to test model privacy. We introduce three tools to make differentially private machine learning more practical: (1) simple sanity checks which can be carried out in a centralized manner before training, (2) an adaptive clipping bound to reduce the effective number of tuneable privacy parameters, and (3) we show that large-batch training improves model performance.
\end{abstract}

\section{Introduction}
Training machine learning models on user data without violating privacy is a challenging problem. One common solution is \emph{differential privacy} \cite{differential-privacy} (DP), a mathematical framework that bounds the contribution of individual entries to some statistic over a database. We identify three practical problems with current approaches to differentially private machine learning:

\begin{enumerate}
\item Privacy guarantees are difficult to interpret. Commonly, training is constrained by parameters $\epsilon$ and $\delta$. These values are difficult to translate into practical guarantees. 
\item In non-private training, multiple models are often trained to find the optimal hyperparameters. In private training privacy spending accumulates for every trained model \cite{chaudhuri2013stability,lei2018differentially}. Additional hyperparameters impact our privacy budget, and should be eliminated where possible.

\item Earlier research treats hyperparameters and privacy parameters as independent \cite{recdp, userdp}. However, this is not always the case. For instance, the batch size influences privacy guarantees, but to preserve model performance, the learning rate must be changed accordingly. 
\end{enumerate}

Most existing work focuses either on preventing private information extraction while reaching acceptable performance, or on optimizing privacy guarantees while reaching performance similar to non-private learning. Neither approach achieves what is desirable in practice: \emph{first determine how much privacy is required for a given task. Then, within this privacy budget, optimize the parameters and hyperparameters of the model}. We offer three methods towards making such a workflow practical.

\paragraph{Preliminaries} A common way to apply DP to deep learning, is through the DPSGD algorithm \cite{recdp}. This algorithm has two important parameters: a \emph{noise scale} $\sigma$ and a \emph{clipping bound} $C$. During gradient descent, any gradient whose $\ell_2$-norm exceeds $C$ is scaled such that its norm is $C$, and Gaussian noise with a variance of $\sigma C$ is added to the gradient. DPSGD can be combined with the \emph{moments accountant} \cite{recdp}, which tracks and controls privacy spending such that it stays within the privacy budget defined by parameters $\epsilon$ and $\delta$ of the DP framework. Privacy-sensitive data often contains multiple records per user, with user identity the sensitive attribute. In such settings, training is often \emph{federated} over the users: each computes a gradient update over their data, applies noise, and the gradients over all data are aggregated centrally. The DP-FedAvg framework \cite{userdp} is a popular extension of DPSGD to the federated setting. 





\section{Methods and experiments}

We perform three experiments, each intended to improve a different part of the DP learning workflow. Section~\ref{section:conclusion} discusses how these are combined into a practical approach to private deep learning.


\subsection{Memorization under differential privacy}

In \cite{understandingrequires}, it was shown that deep neural networks can easily memorize training labels in image classification, even on randomly labeled data. Under differential privacy, such memorization should not be possible.\footnotemark~This allows us to calibrate our privacy parameters: if the model is able to learn a randomly labeled task, the privacy parameters are insufficiently strict.

\footnotetext{The idea that ``differential privacy implies generalization'' is considered folklore \cite{wang2016average}.}

\begin{figure}[t]
\caption{Memorization tests on random noise (a) and on CIFAR-10 (b, c)}
\label{figure:exp1}
\centering
\begin{minipage}{.33\textwidth}
  \centering
  \includegraphics[scale=0.130]{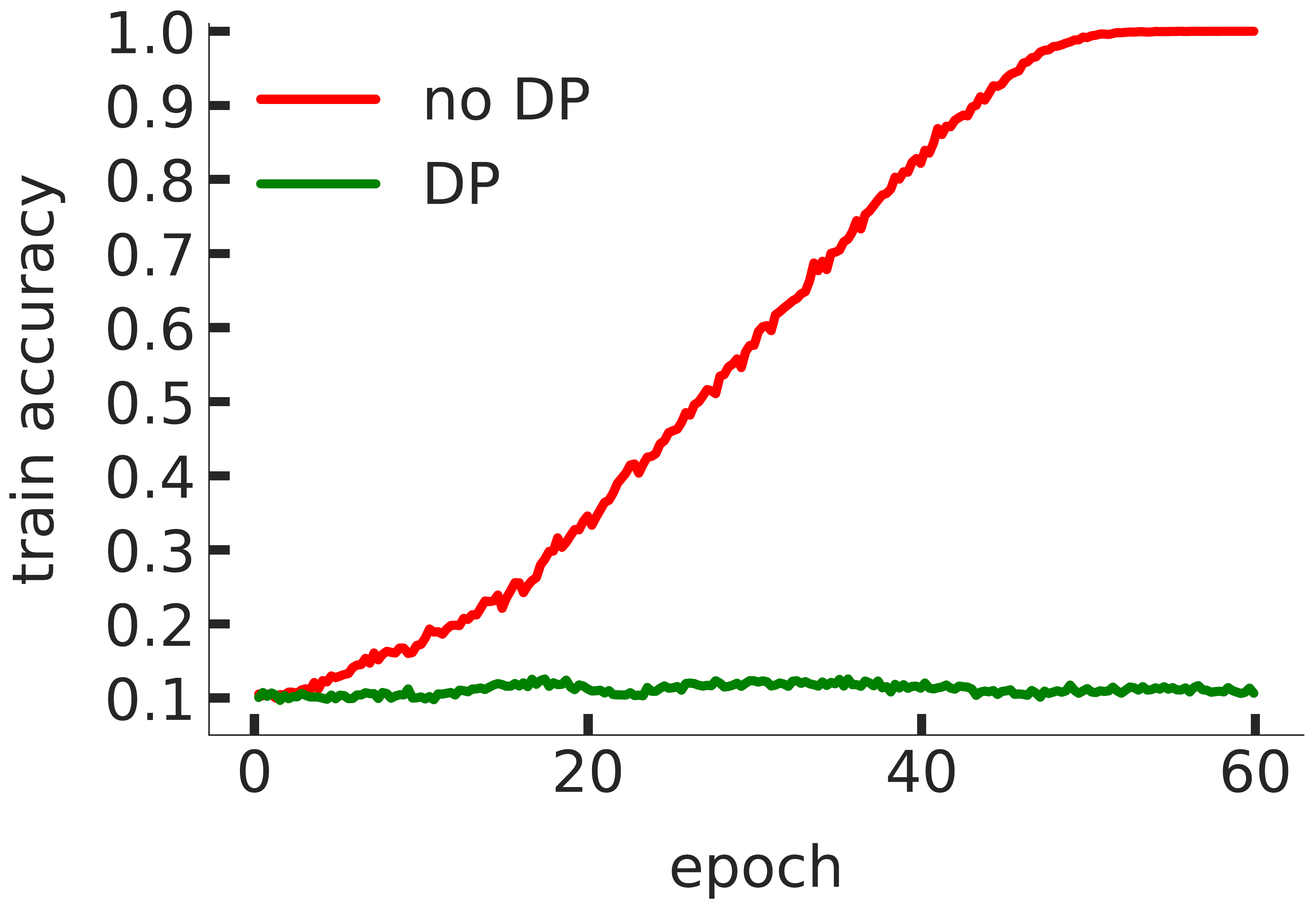}
  \label{fig:random_noise_dp_vs_nodp}
  \caption*{a) Train accuracy}

\end{minipage}%
\begin{minipage}{.33\textwidth}
  \centering
  \includegraphics[scale=0.130]{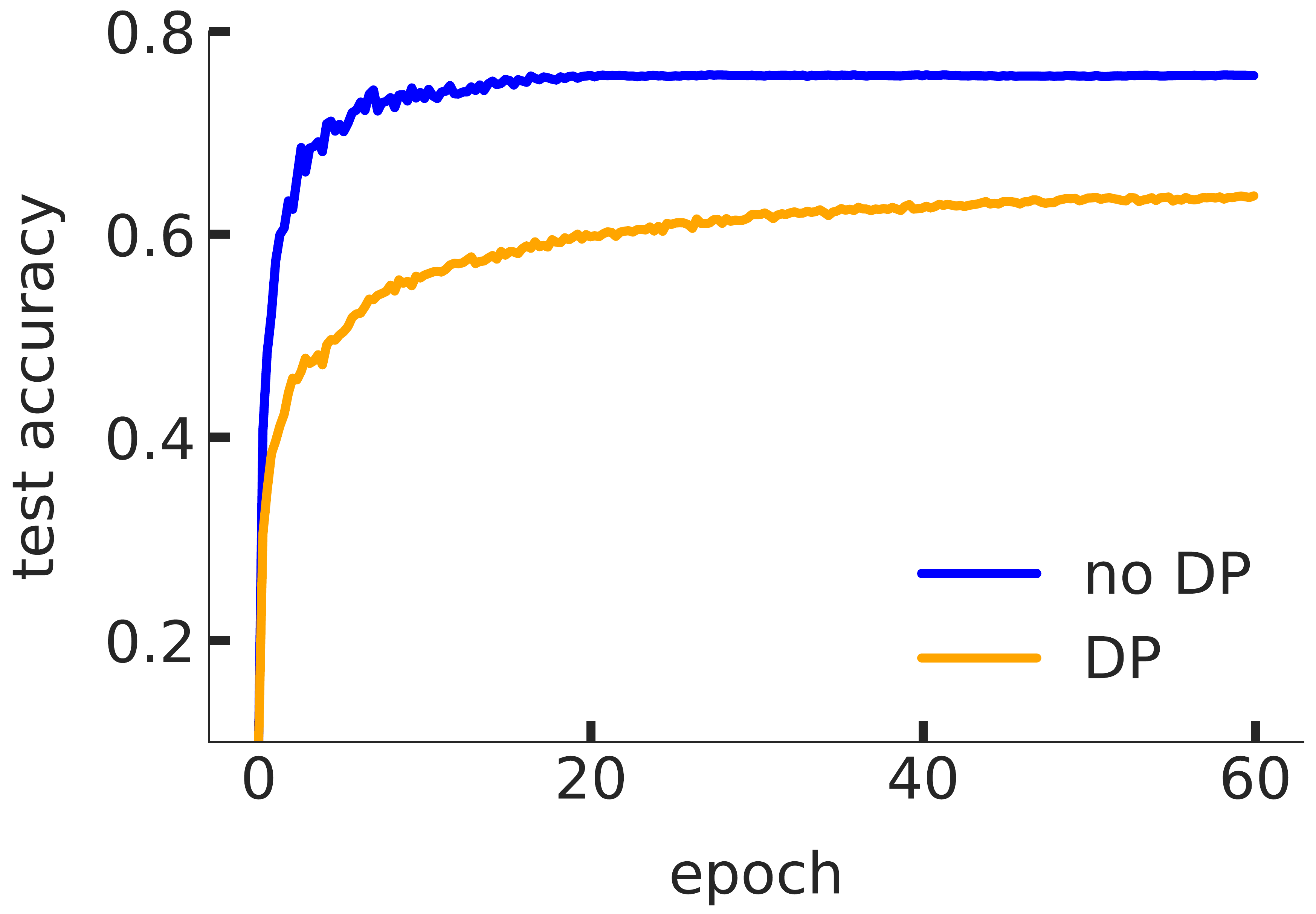}
  \label{fig:test_accuracy_cifar_exp1}
  \caption*{b) Test accuracy}

\end{minipage}
\begin{minipage}{.33\textwidth}
  \centering
  \includegraphics[scale=0.130]{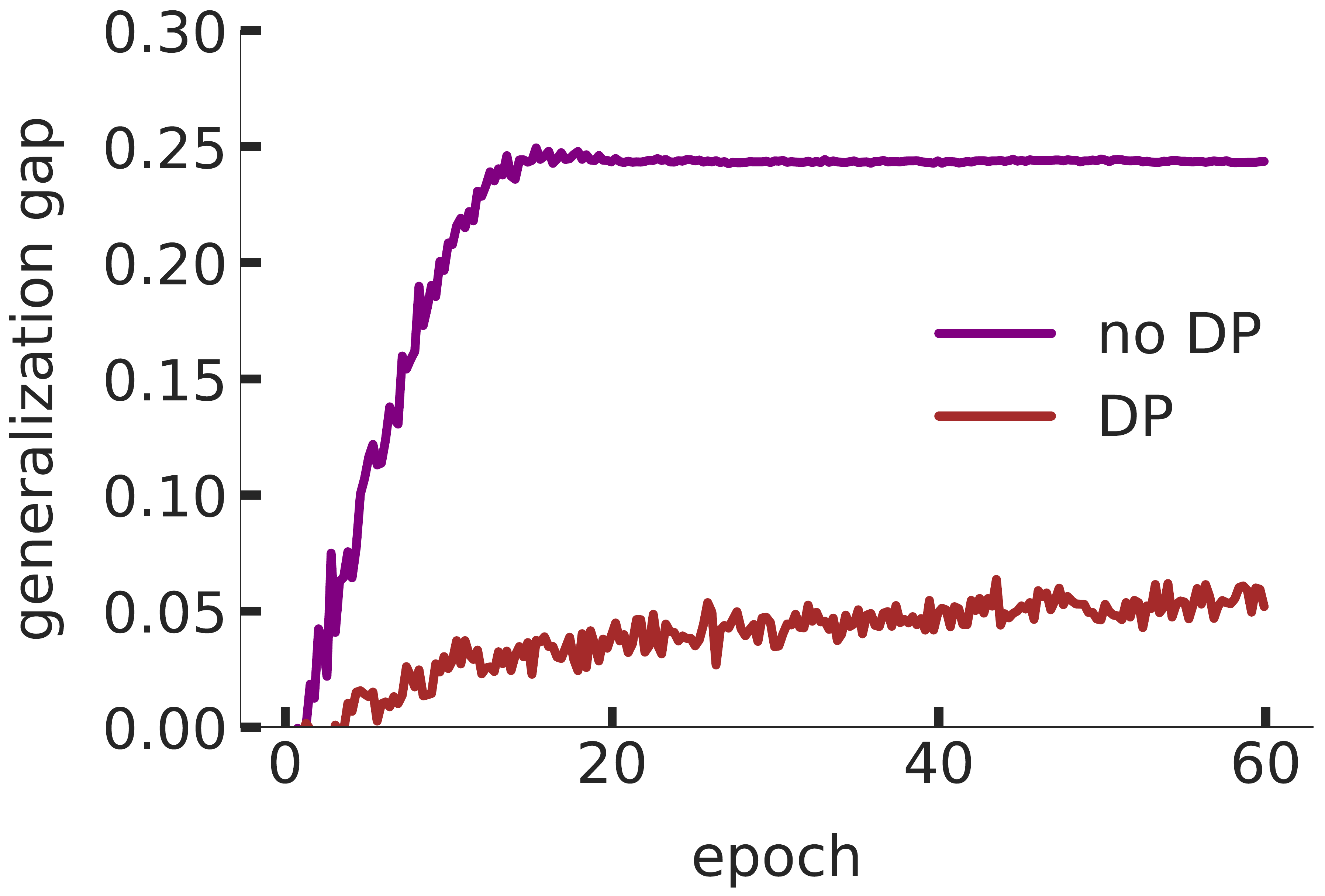}
  \label{fig:generalization gap}
  \caption*{c) Generalization gap}
\end{minipage}
\end{figure}

We train the small Alexnet architecture from Zhang et al. \cite{understandingrequires} for 60 epochs on a dataset of 50,000 random noise examples. We train two models: one with the DPSGD algorithm, extended with momentum and one with non-private SGD with momentum. In the first experiment the models are trained on random noise and the training accuracy is reported. In the second experiment, the same models are trained on the CIFAR-10 task. Here, the test accuracy and difference between train and test accuracy are reported for both models. The differentially private models use $\sigma=0.7225$ for each layer and a batch size of 128, resulting in $\epsilon=20$ for $\delta=\frac{1}{N^{1.1}}$ after 60 epochs. Layers are clipped independently using a clipping bound $C=2.0$. The results are reported in Figure \ref{figure:exp1}.

\noindent

Even with a large privacy spending, DP effectively prevents memorization of random noise, while being capable of learning on real data, with a small reduction in performance.

\paragraph{Memorization in user level differential privacy} As noted in the preliminaries, privacy sensitive data often contains many records per user. In this experiment, we will test memorization in a user-level differential privacy setting, using the DP-FedAvg framework.

\begin{figure}[bt]
\caption{Training MLPs with centralized or distributed patterns}
\label{user-level}
\centering
\begin{minipage}[t]{.5\textwidth}
  \centering
  \includegraphics[scale=0.13]{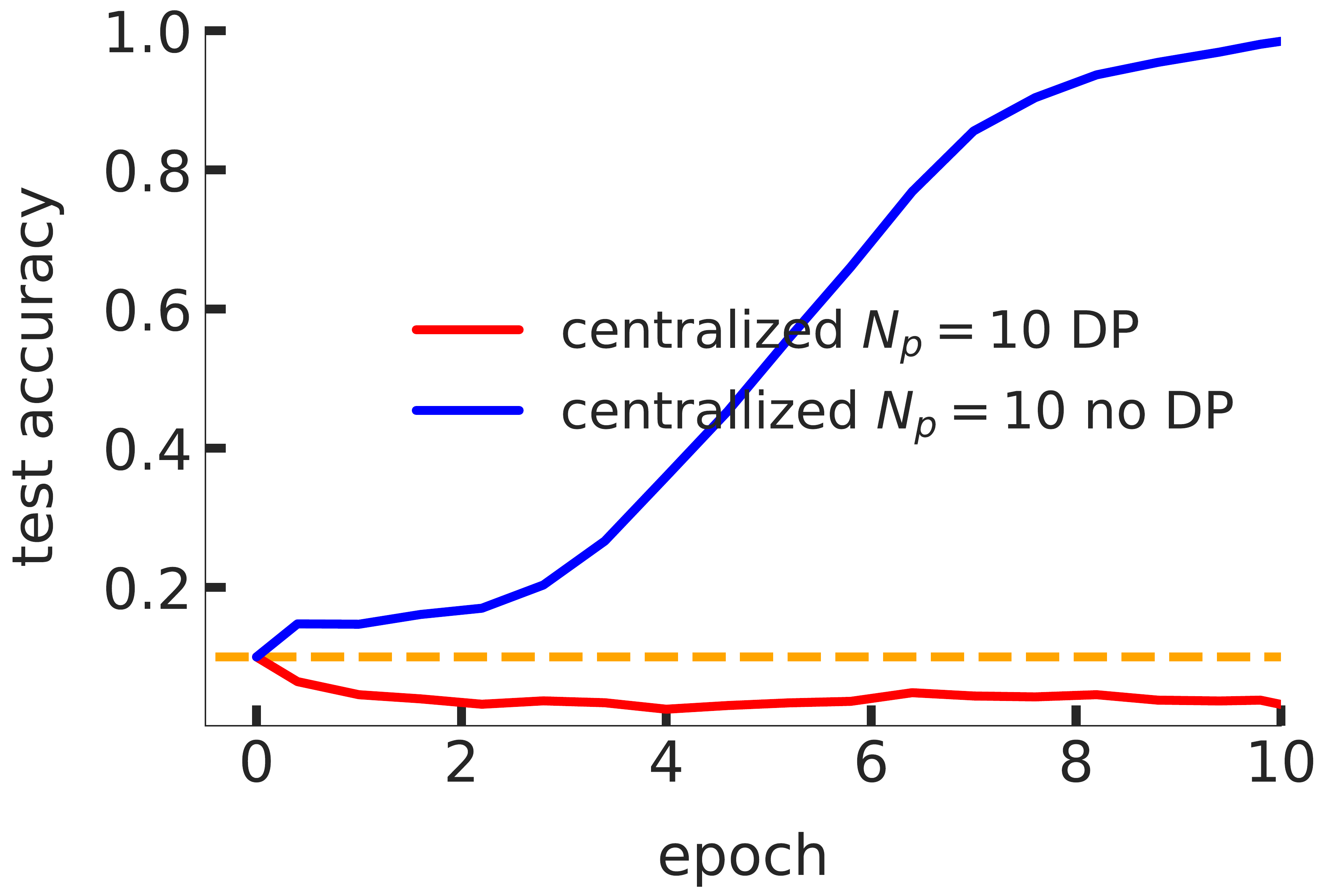}
  \caption*{MLP on centralized dataset}
  \label{1}
\end{minipage}%
\begin{minipage}[t]{.5\textwidth}
  \centering
  \includegraphics[scale=0.13]{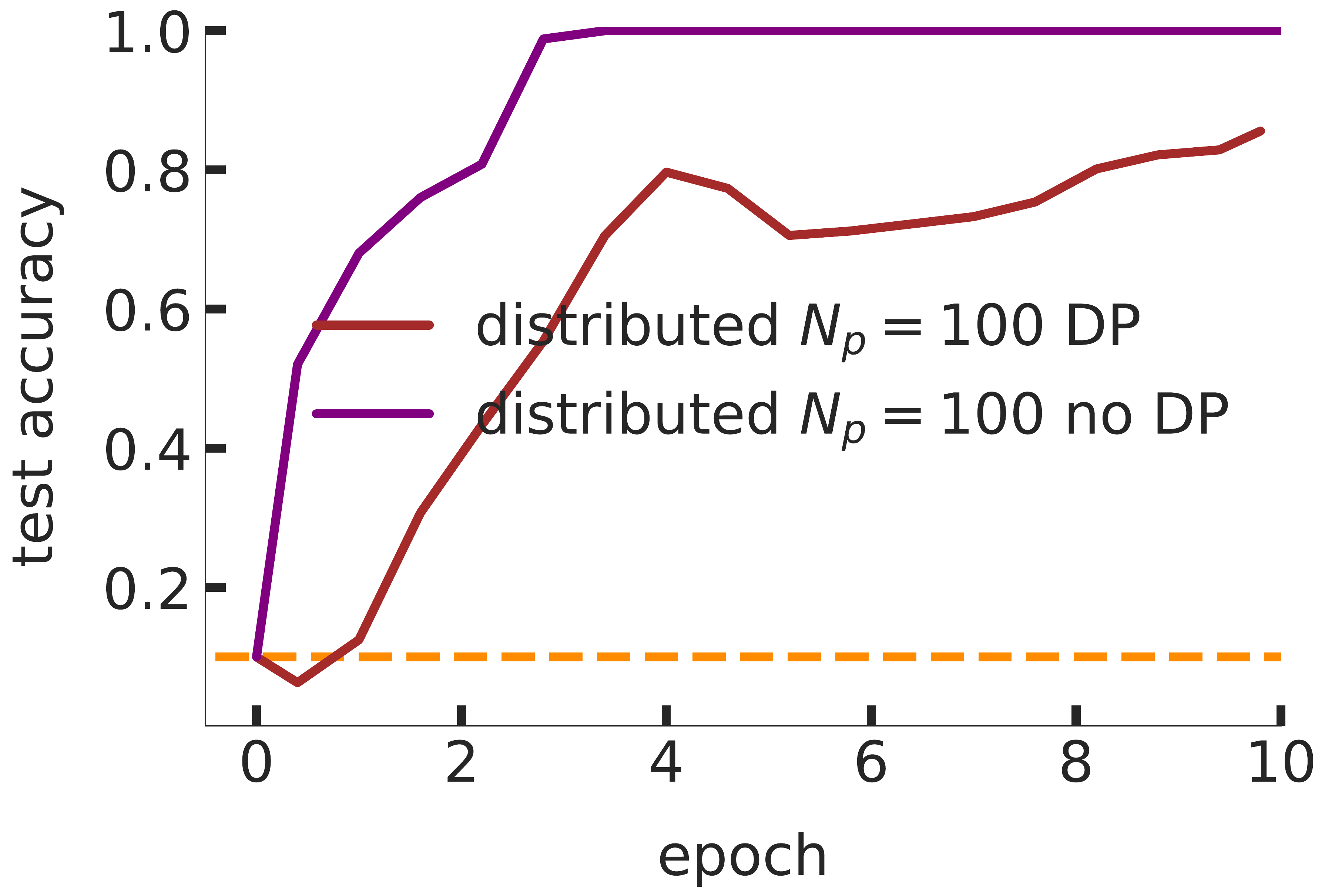}
  \caption*{MLP on distributed dataset $N_p=100$}
  \label{2}
\end{minipage}
\end{figure}

We generate two 10-class datasets of 1,000 users with 10 records each, resulting in 10,000 records. All records contain 28$\times$28 random noise images. All records of one user are assigned the same (random) label. For $N_p$ of the 10,000 records, a pattern is inserted: in the 14$\times$14 upper left patch, all pixels are set to 1.0 and the label is set to 1. In one, the \emph{centralized} dataset, the pattern is inserted only into the records of one user, in the other, the \emph{distributed}, it is inserted uniformly over all records. Figure~\ref{user-level} shows the results. For more details, see \cite{apracapproach}. As expected, in the centralized setting, we are able to learn the pattern without differential privacy but not with differential privacy. 

Distributed patterns \emph{should} be learned. Using the proportions of the centralized setting, DP is too constrictive, but when we increase the occurrence of the pattern, we see that learning is possible. 

\subsection{Adaptive clipping}
\label{section:adaptive-clipping}

\begin{figure}[H]
\begin{minipage}{.5\textwidth}
  \centering
  \includegraphics[scale=0.18]{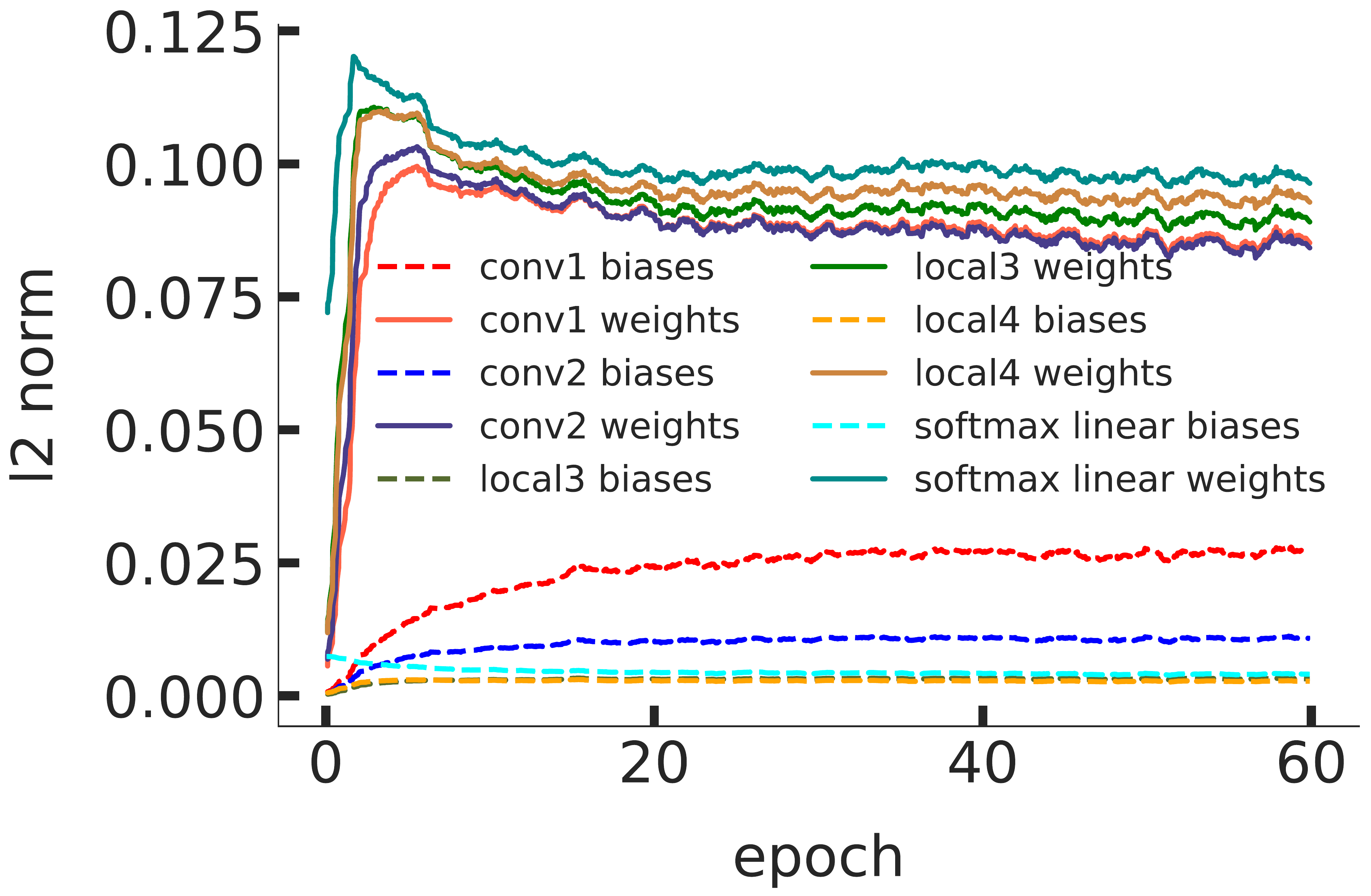}
  \caption{$\ell_2$ norms}
  \label{fig:l2norms_per_layer}
\end{minipage}
\begin{minipage}{.5\textwidth}
  \centering
  \includegraphics[scale=0.18]{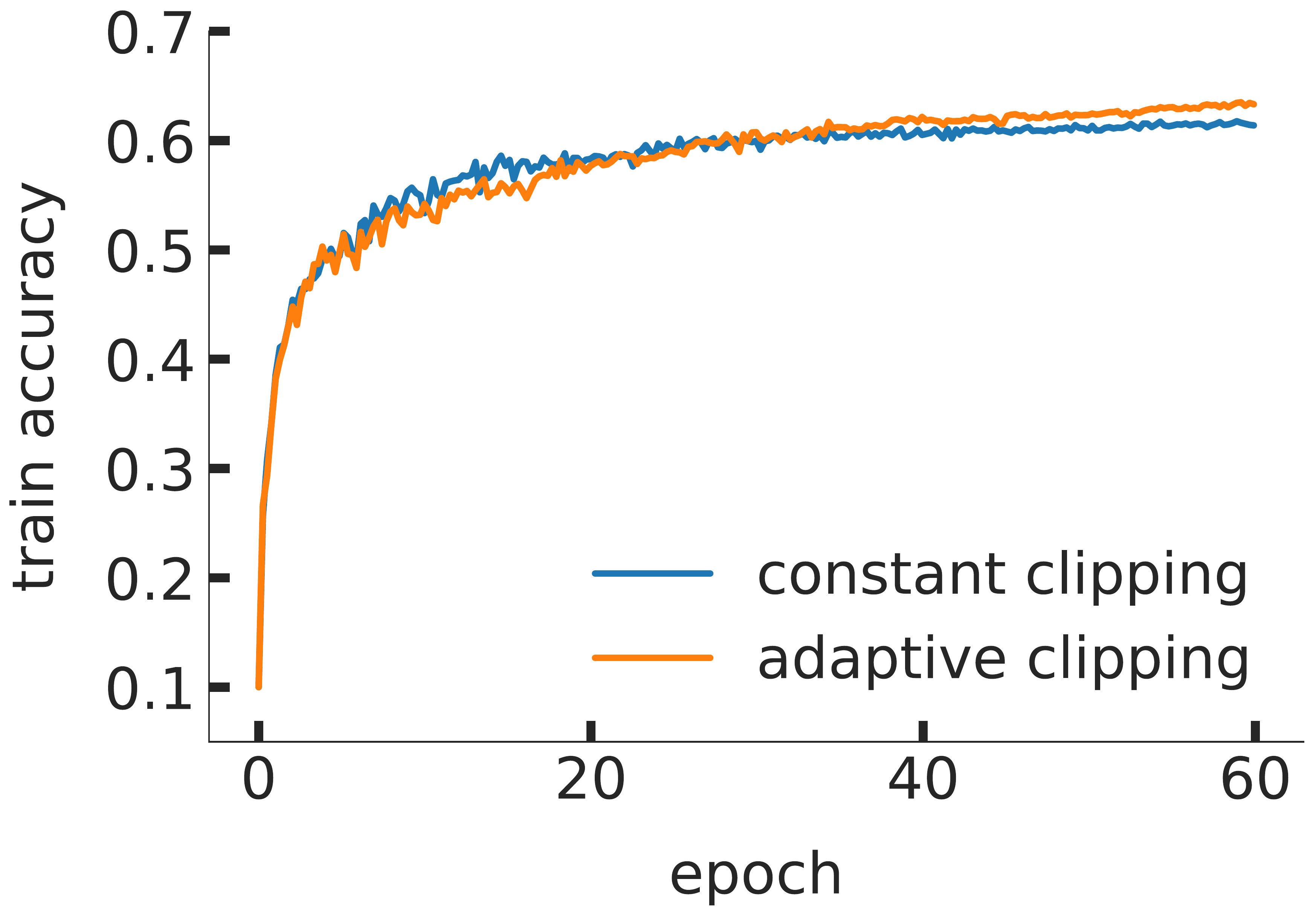}
  \caption{Test accuracy vs clipping methods}
  \label{fig:test_accuracy_cifar}
\end{minipage}
\end{figure}

One of the main challenges for training with DPSGD is choosing a good clipping parameter $C$. To illustrate the difficulty, Figure~\ref{fig:l2norms_per_layer} shows the $\ell_2$-norm of the gradient per layer over the course of training for a non-private model. Two observations can be made. First, the $\ell_2$-norms of layers may be very different in the beginning of training compared to the end of training. Second, the size of the gradient may differ between layers, and between weights and biases. 

Combining these two observations, we propose a gradient-aware clipping scheme. This adaptive clipping schedule uses \emph{the differentially private mean $\ell_2$-norm of the previous batch times a constant factor $\alpha$} as the $\ell_2$ norm bound for the current batch $L$. We define the per layer clipping bound $C_{t}^l$ for round $t$ and layer $l$ over the individual gradients from that layer of the previous round $g_{t-1}^l(x_i)$ and privacy parameters $\sigma_{l^2}$ and $C^l_{l^2t}$: 
\noindent\begin{minipage}{.54\linewidth}
\begin{equation*}
  C_{t}^l = \alpha |L|^{-1}\left( \sum_{i \in L_t} \text{clip}(\lVert g_{t-1}^l(x_i)\lVert_2\right)  + \mathcal{N}(0,\sigma_{l^2}^2 {C_{l^2 t}^l}^2))
  \end{equation*}
\end{minipage}%
\begin{minipage}{.5\linewidth}
  \begin{equation*}
    \text{clip}(\lVert y \lVert_2)= \lVert y \lVert_2 / \max\left(1,\frac{\lVert y \lVert_2}{C^l_{l^2 t}}\right)
  \end{equation*}
\end{minipage}
To choose $C^l_{l^2t}$, we use a similar adaptive procedure: we use the differential private $\ell_2$-norm from the previous iteration $C^l_{t-1}$ times a constant $\beta$:
$C^l_{l^2t} = \beta C^l_{t-1}$.
We choose $\beta=2$ and initialize $C^l_{l^20}$ by training for one iteration on random noise and extracting the mean $\ell_2$-norm. 

We train two versions of the small Alexnet model with the DPSGD algorithm, this time without momentum. The first uses a constant clipping bound, optimized by a grid search over $C$ \cite{apracapproach}. The second uses the adaptive clipping scheme with $\alpha=1.1$. For the adaptive model, we increase the gradient noise scale $\sigma$ to $0.725$ and use an $\ell_2$-norm noise scale $\sigma_{l^2}=2.5$ to use the same privacy budget as the non adaptive model. The test accuracy is reported for both models in Figure \ref{fig:test_accuracy_cifar}. With adaptive clipping the accuracy climbs from 61.6\% to 63.5\%.

\noindent

We replaced parameter $C$ with $\alpha$, $\beta$ and $\sigma_{l^2}$. However, the model is very robust to changes in $\beta$ and $\sigma_{l^2}$. Doubling the values of either $\beta$ or $\sigma_{l^2}$ yields very similar test accuracy. Because our approach is \emph{adaptive}, we expect a single parameter value for $\alpha$ leads to good performance across tasks. 


To examine this hypothesis, we carried out additional experiments on MNIST, CIFAR-10 and CIFAR-100 (reported in \cite{apracapproach}). An $\alpha$ value of $1.0$ is near the optimum for all datasets. This suggests that adaptive clipping results in parameters that are easier to choose without seeing the data. A broader investigation across datasets is required to test this hypothesis further.

\subsection{Large batch training}

\begin{figure}[H]
\centering
\begin{minipage}[t]{.55\textwidth}
  \centering
    \captionof{figure}{$\sigma$/example vs batch size}
    \includegraphics[scale=0.12]{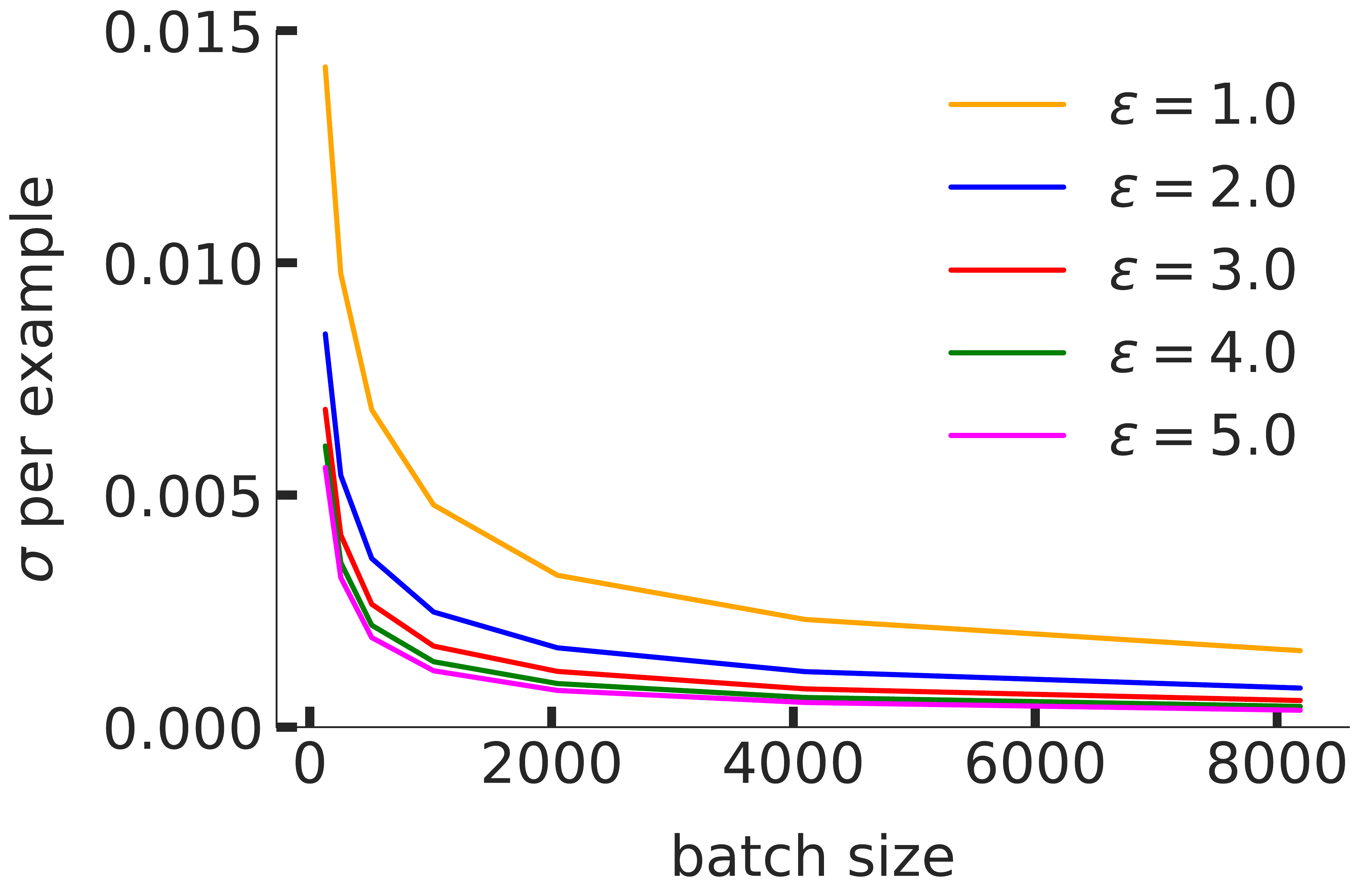}
  \label{fig:sigma_vs_batchsize}
\end{minipage}%
\begin{minipage}[t]{.45\textwidth}
  \centering
  \captionof{table}{Batch size versus accuracy}
   \label{table:batch_size_vs_accuracy_tab}
  \begin{table}[H]
    \centering
    \begin{tabular}{ll}
    Batch size         & accuracy \\
    \hline
    128                & 61.6\%   \\
    512                & 64.2\%   \\
    1024               & \textbf{66.9}\%   \\
    1024 (base lr) & 47.2\%
    \end{tabular}
\end{table}
\end{minipage}
\end{figure}

Training with larger batches reduces privacy spending. To illustrate, Figure~\ref{fig:sigma_vs_batchsize} shows the noise added per example as a function of batch size (using the the moments accountant from \cite{recdp} to find the minimum $\sigma$ that fully uses a pre-defined privacy budget in 10 epochs for a training set of 60,000 examples). Large batches dramatically reduce added noise. However, large batches can strongly hurt performance. In Goyal et al. \cite{largebatchtraining}, several methods are introduced to improve the performance of large-batch training. We adopt the simplest: scaling the learning rate along with the batch size.

We train small Alexnet models with varying batch sizes on CIFAR-10, using the DPSGD algorithm for 60 epochs with a budget of $\epsilon=20$. A base learning rate of 0.01 is used for a batch size of 128. We increase both the the batch size and learning rate by a factor of $k$, and repeat the experiment. We train each model for 60 epochs, until an accumulated privacy loss of $\epsilon=20$ $\pm 0.05$ after 60 epochs is found. Table \ref{table:batch_size_vs_accuracy_tab} shows the results: training DP models with larger batches can be beneficial, but only when the learning rate is scaled accordingly.

\section{Conclusion}
\label{section:conclusion}

For differentially private learning, hyperparameter optimization on sensitive datasets is undesirable. The proposed methods enable an approach to differentially private learning with reduced privacy spending on hyperparameter tuning before training the final model. Given a classification task and data dimensions, we suggest the following approach to choosing the differential privacy parameters:

\begin{itemize}
  \item Choose a model that is successfully tested on a non-private, similar benchmark tasks and use default hyperparameters.
  \item Choose the largest batch size that fits in memory on the training device(s) and scale the learning rate accordingly.
  \item Calibrate the noise scale parameter with the DPSGD \cite{recdp} or DP-FedAvg \cite{userdp} model on a centralized dataset of random noise until the sanity checks succeed.
  \item When all sanity checks have passed, train on private data with the same budget. Use a small portion of the budget for computing the differentially private mean $\ell_2$-norm and use the adaptive clipping method of Section~\ref{section:adaptive-clipping} with default values $\alpha=1.0$, $\beta=2.0$.
\end{itemize}
\noindent

These steps combined, provide the structure of a basic differential privacy training workflow. This is far from a full-proof approach: the sanity checks function more as unit tests than hard guarantees, some architectures, like conditional models \cite{secretsharer}, are not yet supported, and it is not clear whether this approach is sufficient when adversaries actively attempt to influence the training process, such as in the privacy attacks proposed by Hitaj et al. \cite{dmug}. We hope that our approach provides a basis that can be extended to study such questions.




\paragraph{Acknowledgements} We thank the reviewers for their valuable comments. This publication was supported by the Amsterdam Academic Alliance Data Science (AAA-DS) Program Award to the UvA and VU Universities.

\bibliographystyle{plain}
\bibliography{references}

\end{document}